\documentclass[letterpaper,10pt,conference]{ieeeconf}
\IEEEoverridecommandlockouts
\let\labelindent\relax
\usepackage{float}
\usepackage{tabularx}      % 为了更灵活的表格宽度控制  
\usepackage{makecell}      % 为单元格内容换行  
\usepackage{tcolorbox}
\usepackage{handy_style}
\definecolor{improvegreen}{rgb}{0,0.6,0}  
\definecolor{degradered}{rgb}{0.8,0,0} 
\let\oldnl\nl% Store \nl in \oldnl
\newcommand{\nonl}{\renewcommand{\nl}{\let\nl\oldnl}}% Remove line number for one line
\newcolumntype{x}{>{\columncolor{MistyRose}}c}
\newcolumntype{y}{>{\columncolor{LightCyan1}}c}

\newcommand\blfootnote[1]{%
  \begingroup
  \renewcommand\thefootnote{}\footnote{#1}%
  \addtocounter{footnote}{-1}%
  \endgroup
}

\acrodef{urdf}[URDF]{Unified Robot Description Format}

\title{\LARGE \bf Gold Points Sniper: Self-guided Visual Reasoning in VLM \\for Fine-grained Action Understanding}

\author{Haodi Liu$^{1}$, Xinhang Yang$^{1}$, Kunda Yan$^{1}$, Sen Cui$^{1}$, Zeyu Zhang$^{2}\dagger$, Changshui Zhang$^{1}\dagger$% <-this % stops a space
}

\begin{document}

%%%%%%%%%%%%%%%%%%%%%%%%%%%%%%%%%%%%%%%
% Journal headline
%%%%%%%%%%%%%%%%%%%%%%%%%%%%%%%%%%%%%%%
% \markboth{IEEE Robotics and Automation Letters. Preprint Version. Accepted MAY, 2025}
% {Zhang \MakeLowercase{\textit{et al.}}: M3Bench: Benchmarking Whole-body Motion Generation for Mobile
% Manipulation in 3D Scenes}

\maketitle
\blfootnote{%
    $^{1}$ Beijing National Research Center for Information Science and Technology (BNRist), Department of Automation, Tsinghua University, Beijing, P.R.China 

    $^{2}$ State Key Laboratory of General Artificial Intelligence, Beijing Institute for General Artificial Intelligence (BIGAI).
}
\blfootnote{$\dagger$ Corresponding authors. Zeyu Zhang <zhangzeyu@bigai.ai>, Changshui Zhang <zcs@mail.tsinghua.edu.cn>}
% \vspace{-2cm}

\begin{abstract}
Robots operating in everyday environments must understand fine-grained human actions, intentions, and contextual cues from broad views where people occupy only small regions, a capability unmet by current systems.
While open-vocabulary action recognition methods remain limited to assigning predefined labels, and vision-language models (VLMs) face an inherent trade-off between informational richness and factual fidelity in their outputs, neither approach achieves the deep semantic interpretation required for reliable human-robot interaction. We propose Gold Points Sniper (GPS), a novel framework that empowers lightweight VLMs with self-guided multimodal reasoning capabilities for fine-grained human action understanding. Our approach comprises three key modules: Gold Points Extractor trains VLMs to identify critical action-relevant details, Selective Socratic Questioner validates and refines these details through selective self-questioning, and Semantic Entailment Evaluator quantitatively assesses factual consistency using semantic entailment classification. Extensive experiments on our curated instruction-tuning dataset based on the CAP benchmark demonstrate that GPS-enhanced lightweight VLMs achieve substantial performance improvements, with some models reaching performance comparable to proprietary GPT-4o while maintaining superior factual accuracy. Our work establishes a reliable foundation for fine-grained action understanding in domestic robotics, enabling robots to safely interpret human behavior through information-dense yet factually grounded descriptions.  Source code, training configurations, annotation prompts, and dataset details are released at \url{https://github.com/Haodi-Liu/GPS-Gold-Point-Sniper}.
% The effective integration of robots into human environments demands a nuanced understanding of human actions, a task where current Vision-Language Models (VLMs) often struggle with a trade-off between descriptive richness and factual accuracy. While large-scale models demonstrate impressive capabilities, their significant computational requirements hinder practical deployment on resource-constrained platforms like robots. To bridge this gap, we propose the \textbf{Gold Points Sniper (GPS)}, a framework that empowers lightweight VLMs to perform self-guided multimodal reasoning for fine-grained human action understanding. GPS introduces "Gold Points"—core, verifiable factual details extracted from visual input—to anchor the model's reasoning process. The framework is composed of three key modules: (i) a \textbf{Gold Points Extractor} to establish a factual foundation, (ii) a \textbf{Selective Socratic Questioner} to validate and selectively refine these facts, mitigating error accumulation, and (iii) a \textbf{Semantic Entailment Evaluator} for robust assessment of output factuality. We construct a comprehensive instruction-tuning dataset based on the CAP corpus to train our framework. Extensive experiments show that GPS endows lightweight VLMs with the capability to comprehensively and accurately describe human action, associated intention, and context. This presents a crucial advancement, enabling the deployment of highly perceptive yet efficient models on embodied agents.
\end{abstract}
% \begin{IEEEkeywords}
% Fine-grained Human Action Understanding, Vision-Language Models, Self-Guided Reasoning, Embodied AI
% \end{IEEEkeywords}

\section{Introduction}
%%%%%%%%%%%%%%%%%%%%%%%%%%%%%%%%%%%%%%%
% Intro outline
%%%%%%%%%%%%%%%%%%%%%%%%%%%%%%%%%%%%%%%
% > The logic flow of the introduction section is listed as below, write down the introduction by following step 1 through 9.

% 1. Begin with the background (the scope) and motivation of this project.
% 2. Summarize the limitation of existing work regarding the problem and why they have such limitations.
% 3. Propose your research problem in a concise, specific, logical, and promising way.
% 4. After 3, describe the consequence of leaving this problem unsolved. (directly link the limitations of existing work with the consequences)
% 5. Point out the essential challenges of this problem.
% 6. To tackles these challenges, describe the capabilities of the proposed model and provide a concise overview of its mechanism.
% 7. Explain how your approach tackles identified challenges.
% 8. Briefly mention the kind of evidence or results that back up the performance of the model, adding credibility.
% 9. Highlight the core contribution of your work, and briefly indicate how it advances or impacts the field.

\begin{figure}[t]
\centering
\includegraphics[width=\linewidth]{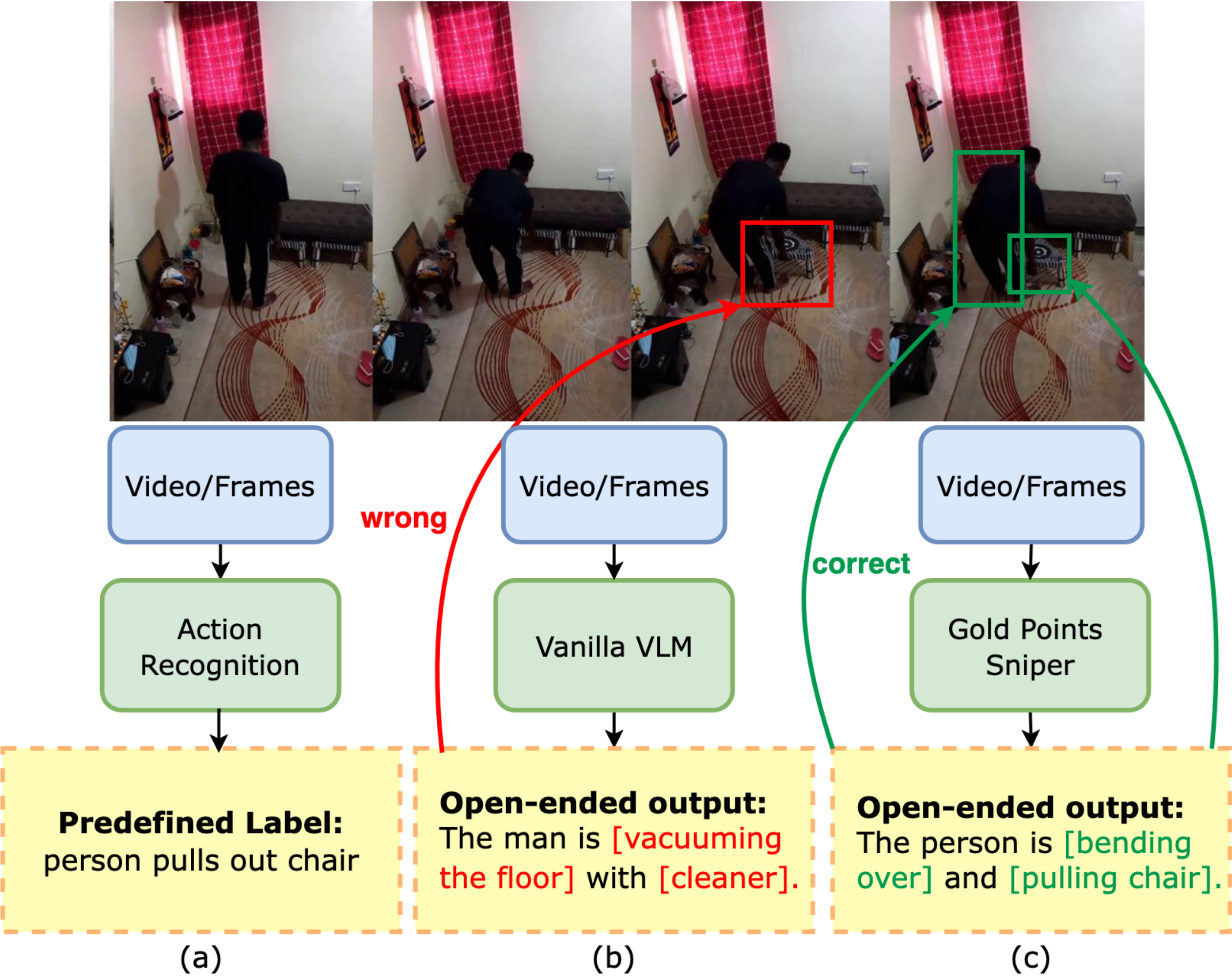}
\caption{The image on top is an example video frames from CAP\cite{byrne2023fine} with label "person pulls out chair". (a). Action recognition gives precise but limited labels. (b). VLMs' outputs are informative but error-prone (shown in red). (c). Our \textbf{GPS} framework enables VLMs to precisely capture multiple fine-grained action clues (in green).} 
\label{fig:intro}
\end{figure}

The increasing adoption of robots in domestic environments necessitates a deeper understanding of human behavior. For robots to effectively collaborate, assist, and socialize with people, the ability to accurately recognize and interpret human actions is essential. This capability extends beyond merely identifying predefined activities from cropped, human-centric images; it requires sophisticated understanding of human actions, intentions, and contextual cues in broader views where humans occupy only a small portion of the image (see \cref{fig:intro}~top). However, neither open-vocabulary action recognition \cite{chen2023video, zhang2024enhancing, cheng2024denoiser, jia2024generating, chen2023large, huang2024froster, lu2024enhancing, yu2025learning} nor vision-language models (VLMs) \cite{bai2023qwenvlversatilevisionlanguagemodel, liu2024improved, chen2023minigpt, hurst2024gpt, li2024llavanext-strong, lillava} fully achieves this level of deep semantic interpretation, as each approach suffers from inherent shortcomings.

Despite technical advancements in open-vocabulary action recognition, including cross-modal alignment \cite{chen2023video, zhang2024enhancing, cheng2024denoiser}, knowledge augmentation \cite{jia2024generating, chen2023large}, and generalizable learning \cite{huang2024froster, lu2024enhancing, yu2025learning}, current methods remain fundamentally limited to assigning predefined action labels \cite{kay2017kinetics, caba2015activitynet, kuehne2011hmdb, soomro2012ucf101} to visual inputs, as illustrated in \cref{fig:intro}~(a).
These approaches fail to capture details about individuals, their environment, and the nuances of their interactions, thereby precluding deeper semantic interpretation and reasoning.
While VLMs offer a promising alternative with their strong expressive capability for open-ended generation, they face an inherent trade-off between informational richness and factual fidelity: succinct descriptions often omit critical details, while verbose ones become prone to redundancy and hallucinations, as demonstrated in \cref{fig:intro}~(b). 
Although recent work has attempted to address these issues through structured reasoning\cite{wei2022chain, zhang2024multimodal, he2024multi, mondal2024kam, zhang2023cumulative} and self-questioning\cite{zhang2024visual, chen2025putting, sun2024sq, hu2025socratic}, this fundamental trade-off remains a significant limitation.

In this work, we aim to empower VLMs with self-guided generation capabilities for fine-grained human action understanding. 
This approach ensures outputs that distill core, task-relevant information into summaries that are both information-dense and factually accurate, as illustrated in \cref{fig:intro}~(c). 
Our objective is to enable VLMs to generate grounded and insightful summaries, capturing key attributes of people and scenes while elucidating the details of their interactions.
However, achieving this goal is hindered by three major challenges: (i) Lightweight VLMs often exhibit limited visual grounding capabilities, making it challenging to accurately identify and extract key details in response to direct prompts. (ii) VLMs' inherent tendency toward hallucination poses significant risks during self-guided reasoning. Self-generated intermediate steps become contaminated, causing errors to accumulate through the reasoning process and result in fundamentally flawed outputs.
(iii) The lack of appropriate metrics for assessing semantic accuracy and systematic frameworks to distinguish facts from hallucinations creates significant evaluation challenges for open-ended VLM outputs, hindering reliable assessment essential for trustworthy deployment.

To address these challenges, we propose \textbf{Gold Points Sniper (GPS)}, a framework that trains VLMs to self-guide multimodal reasoning and evaluates open-ended outputs through semantic entailment. Gold Points, our core concept, represent fine-grained key information highly relevant to human actions. Our framework comprises three main modules:
(i) The \textbf{Gold Points Extractor} trains VLMs to accurately capture critical, task-relevant details from visual inputs, enhancing visual grounding capabilities. By focusing on grounded core details, VLMs can better guide their subsequent reasoning.
(ii) The \textbf{Selective Socratic Questioner} trains VLMs to verify and refine extracted gold points through selective self-questioning. This internal validation ensures reasoning step integrity, mitigating error compounding and leading to reliable final summaries.
(iii) The \textbf{Semantic Entailment Evaluator} employs semantic entailment classification to quantitatively assess factual consistency between VLM outputs and ground truth details.
To facilitate framework training, we introduce an instruction-tuning dataset based on Consented Activity of People (CAP)~\cite{byrne2023fine} that benchmarks semantic alignment of VLM outputs against fine-grained action understanding.
Our experiments demonstrate that our assessment module aligns closely with human intuition, and the overall framework enables multiple lightweight VLMs to generate more detailed and precise descriptions of human actions.

In summary, our contribution is threefold:
\begin{enumerate}
    \item We propose the Gold Points
Sniper (GPS) framework that enables lightweight VLMs to perform self-guided multimodal reasoning for fine-grained human action understanding, achieving information-dense and factually accurate outputs.
    \item We curate a comprehensive instruction-tuning dataset based on the \textbf{CAP}~\cite{byrne2023fine} that facilitates training and evaluation of semantic alignment between VLM outputs and fine-grained action understanding in vision-language reasoning tasks. 
    \item Through extensive experiments, we demonstrate that our framework enables multiple lightweight VLMs to generate significantly more detailed and precise descriptions of human actions while maintaining factual accuracy.
\end{enumerate}

\section{Related work}

\subsection{Open-Vocabulary Action Recognition}
Recent advances in open-vocabulary action recognition (OVAR) have been driven by CLIP-series models\cite{radford2021learning, ma2022x, rasheed2023fine, zhulanguagebind}, with methodologies categorized into three paradigms.
\textbf{Cross-modal alignment} approaches~\cite{chen2023video, zhang2024enhancing, cheng2024denoiser} have evolved from coarse video-label matching to fine-grained strategies that establish detailed correspondences between visual and textual modalities, while addressing issues like scene bias and noisy labels.
\textbf{Knowledge distillation} methods~\cite{huang2024froster, lu2024enhancing} transfer semantic understanding from powerful VLMs to efficient architectures through teacher-student frameworks and offline knowledge distillation.
\textbf{Prompt engineering} techniques~\cite{jia2024generating, chen2023large} inject domain knowledge by leveraging LLMs to generate enriched textual representations and coordinate multiple model descriptions.
Despite leveraging VLMs' multimodal capabilities, these approaches remain limited to predefined action labels, hindering fine-grained understanding of human activities and their surroundings.
In our work, we propose the GPS framework that exploits VLMs to generate detailed descriptions of actions, intentions, and contextual cues, establishing a reliable and informative perceptual foundation for domestic robots.

\subsection{Structured Reasoning and Self-Questioning}
Extensive research has focused on enabling LLMs and VLMs to generate high-quality responses aligned with user demands through two primary approaches.
\textbf{Structured reasoning}, pioneered by Chain-of-Thought (CoT)~\cite{wei2022chain}, demonstrates that step-by-step reasoning processes enhance LLM performance\cite{zhang2024multimodal, mondal2024kam}.
\textbf{Self-questioning} techniques guide models toward focusing on salient details for more meaningful responses by providing higher-quality questions.
For example, SQ-LLaVA\cite{sun2024sq} and Socratic Questioning\cite{hu2025socratic} generate content-relevant questions about images, while other works\cite{zhang2024visual} refine this by selectively applying self-questioning only to problems requiring in-depth reasoning.
However, these methods face a trade-off between informational richness and factual fidelity.
Our GPS framework integrates reliable knowledge accumulation, intermediate result verification, and self-questioning refinement for fine-grained human action understanding.
Such factual grounding is essential for safe and reliable domestic robots, as inaccurate perception of human behavior can lead to unsafe actions and erode user trust in physical world interactions.

\section{Method}
\label{method}
\begin{figure*}[t]
\centering
\includegraphics[width=\linewidth]{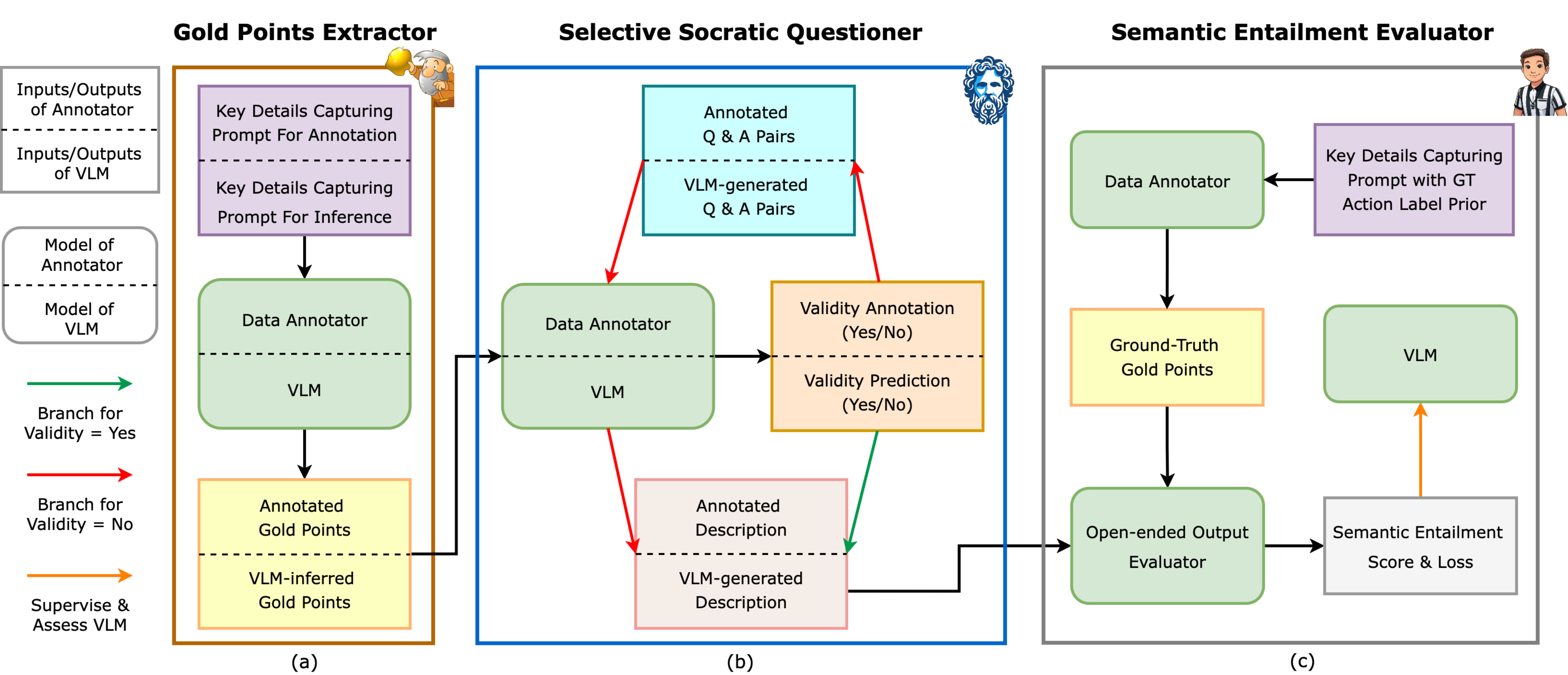}
\caption{Overview of the GPS framework. Dashed blocks are split into an upper section for data annotation and a lower section for VLM inference, with the annotated data used to fine-tune the VLM. Arrows denote data flow, and colored arrows highlight specific relationships. Video frames serve as the implicit input to all Data Annotator and VLM blocks. The Semantic Entailment Evaluator module has no dashed block because annotation and inference workflows diverge over there. The data annotator and output evaluator employed in our work are LLaVA-OneVision-Qwen2-72B-ov-chat\cite{lillava} and LLaMA-3.3-70B-Instruct\cite{grattafiori2024llama} respectively.} 
\label{fig:frame}
\end{figure*}

This section details each module of the GPS framework shown in \cref{fig:frame}. While training and evaluating our framework requires extensive data annotation, a process made costly by manual labor or proprietary APIs, we bypass these bottlenecks by leveraging powerful open-source models to automate data annotation and outputs evaluation. Specifically, we employ LLaVA-OneVision-Qwen2-72B-ov-chat \cite{lillava} for annotation and LLaMA-3.3-70B-Instruct\cite{grattafiori2024llama} for evaluation.

\begin{figure}[h]
\centering  
\includegraphics[width=\linewidth]{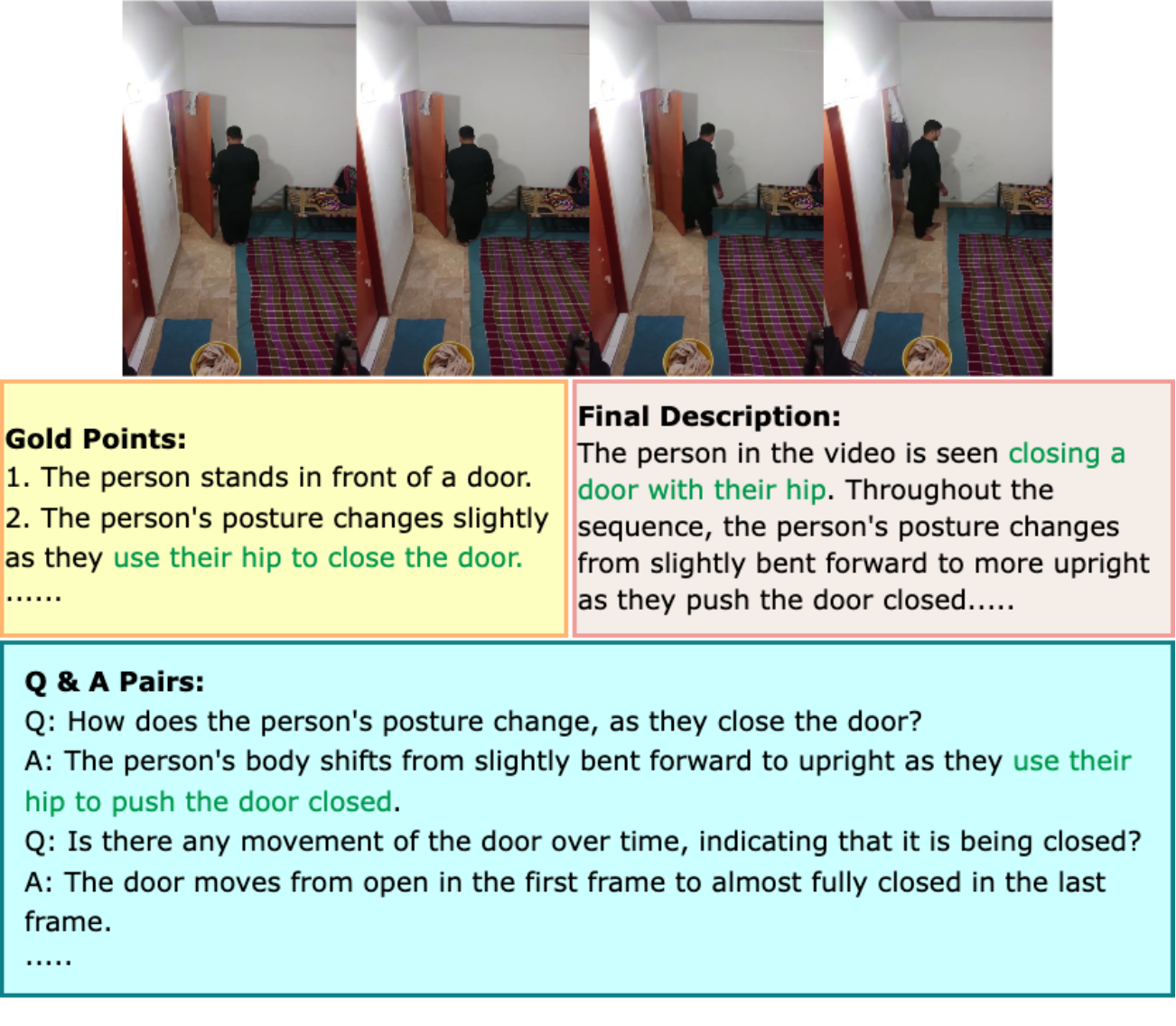}
\caption{Example video frames with GT label "person closes door with hip". Associated gold points, Socratic Q\&A pairs, and final description are provided as demonstrations, with key action information highlighted in green.}
\label{fig:exam}
\end{figure}
\subsection{Gold Points Extractor}

As illustrated in \cref{fig:frame}~(a), the Gold Points Extractor module trains a VLM to identify critical action-relevant details from input video frames and formulate them as Gold Points (see yellow box in \cref{fig:exam} for examples). 

Let $X_v$ denote the input video frames, $\mathcal{G}$ the VLM serving as the Gold Points Extractor, $X_{cap}$ the key details capturing prompt for inference, and $X_{gold}=\{X_{gold,i}\}_{i=1}^N$ a set of $N$ \textbf{VLM-inferred Gold Points} (where $N$ varies by case). The gold points extraction operation is:
\begin{equation}
  X_{gold} = \mathcal{G}(X_{cap}, X_v)
\label{eq:gpe_inf}
\end{equation}
as our gold points extraction operation.

The data pairs used to train $\mathcal{G}$ are $\{X_v, X^{ann}_{gold}\}$, where the \textbf{Annotated Gold Points} $X^{ann}_{gold}$ are generated by:
\begin{equation}
    X^{ann}_{gold}=\mathcal{A}(X^{ann}_{cap}, X_v)
\end{equation}
Here, $\mathcal{A}$ represents the data annotator operating on video frames $X_v$ with annotation prompt $X^{ann}_{cap}$. The training objective $\mathcal{O}_{GPE}$ maximizes the likelihood of generating each annotated gold point $X_{gold,i}^{ann}$ given the visual input $X_v$ and all previously produced points $X_{gold,<i}^{ann}$:
\begin{equation}
  \mathcal{O}_{GPE} = \sum_{i=1}^{N} \log P(X^{ann}_{gold,i} \mid X_v, X_{cap}, X^{ann}_{gold,<i})
\label{eq:gpe_ob}
\end{equation}

Inference prompts (e.g., $X_{cap}$) are concise versions of their annotation counterparts (e.g., $X^{ann}_{cap}$), abbreviated to meet fine-tuning token limits. For simplicity, we henceforth use data notation (e.g., $X_{gold}$) as shorthand for the complete prompt containing it when displayed as model input.

The inferred gold points $X_{gold}$ subsequently guide the VLM's reasoning throughout downstream tasks, serving as foundational reference. In optimal cases, these points capture all critical information about actions, intentions, and context. Even when incomplete, they still provide substantial visual cues that establish a solid foundation for extracting and integrating finer-grained details in following processing stages.

\subsection{Selective Socratic Questioner}

Since $X_{gold}$ may be incomplete or flawed, Selective Socratic Questioner (\cref{fig:frame}~(b)) trains the VLM to validate and selectively refine these points through a two-stage process.

\textbf{Stage 1: Validating Gold Points.} The VLM first assesses whether the gold points are accurate and sufficient for understanding the subject's action, intent, and context. Let $\mathcal{V}$ denote the validator VLM. The validation process is:

Let $\mathcal{V}$ be the VLM playing the role of validator, then the gold points validation process can be formulated as
\begin{equation}
  y = \mathcal{V}(X_v, X_{gold})
\label{eq:ssq_ver}
\end{equation}
where $y$ is a boolean indicator (1 for valid, 0 for invalid).

\begin{figure}[t]
\centering  
\includegraphics[width=\linewidth]{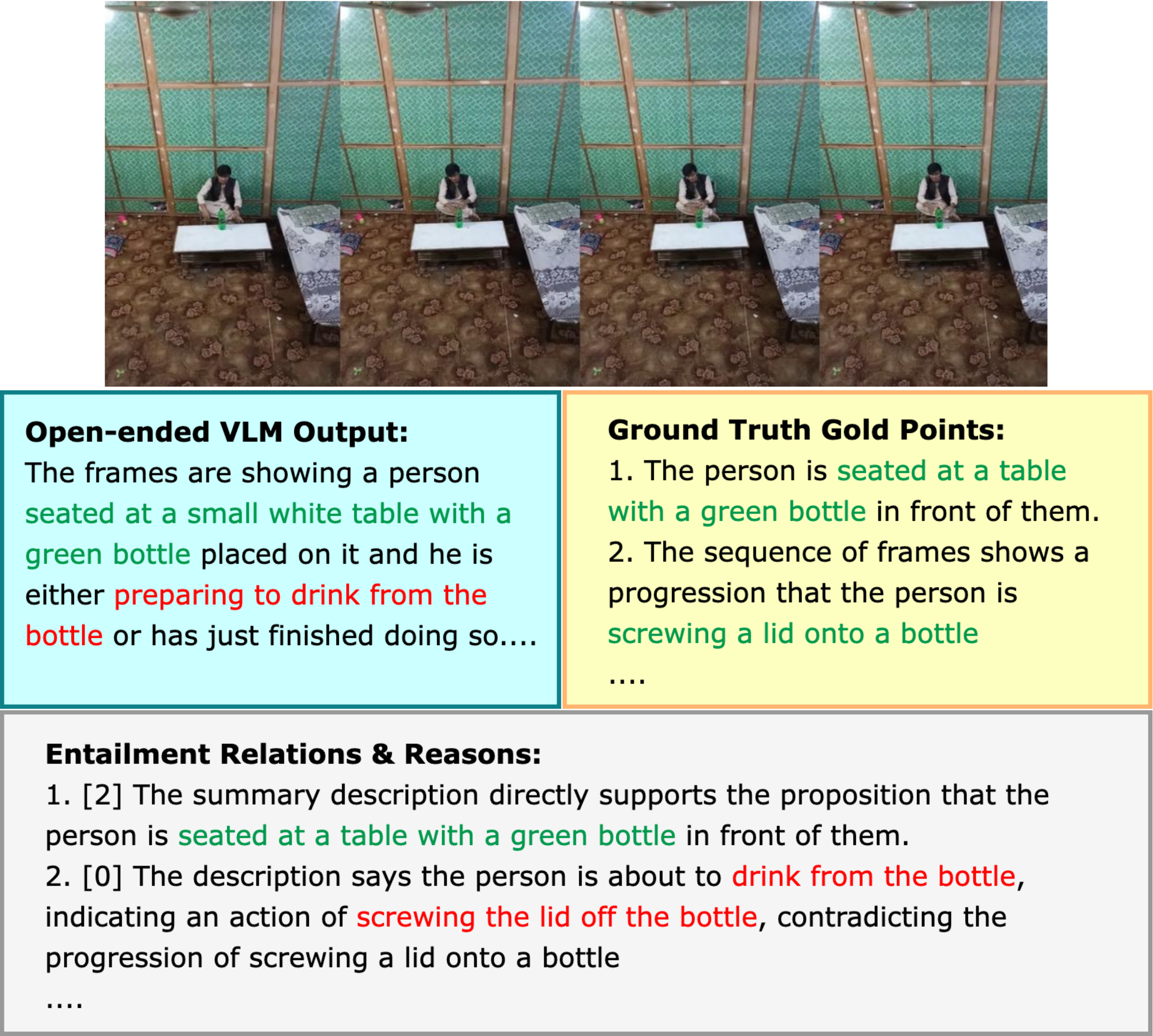}
\caption{Example video frames with GT label ``person screws lid to bottle''. Associated GT gold points, entailment relations (Contradiction 0/Neutrality 1/Entailment 2) with reasons, and VLM output are shown, where entailments of key action information are highlighted in green and contradictions are highlighted in red.}
\label{fig:see}
\end{figure}

To better align with real inference, we initially annotate $X^{ann}_{gold}$ without access to the GT action label $B$. For training $\mathcal{V}$, we then provide the annotator $\mathcal{A}$ with $B$ to determine whether $X^{ann}_{gold}$ alone suffices to robustly predict $B$ from $X_v$:
\begin{equation}
  y^{ann} = \mathcal{A}(X_v, X^{ann}_{gold}, B)
\label{eq:ssq_ann}
\end{equation}
where $y^{ann}$ is the annotated boolean validity. Thereafter, $\{(X_v, X^{ann}_{gold}), y^{ann}\}$ serve as training data for $\mathcal{V}$ and the training objective $\mathcal{O}_{valid}$ is the negation of BCE loss: 
\begin{equation}
    \mathcal{O}_{valid}=-BCE(y,y^{ann})
\end{equation}

\textbf{Stage 2: Selective Questioning Refinement.} The module's workflow branches based on the validity assessment. When $X_{gold}$ are deemed valid ($y=1$), the describer VLM $\mathcal{D}$ generates the final description $X_{des}$ directly:

\begin{equation}
  X_{des} = \mathcal{D}(X_v, X_{gold} \mid y=1) 
\end{equation}

The corresponding training data for $\mathcal{D}$ consists of $\{(X_v, X^{ann}_{gold}), X_{des}^{ann}\}$ where $y^{ann}=1$, with annotated description $X_{des}^{ann}$ obtained via:
\begin{equation}
    X_{des}^{ann}=\mathcal{A}(X_v,X^{ann}_{gold} \mid y^{ann}=1)
\end{equation}

When $X_{gold}$ are deemed invalid ($y=0$), Selective Socratic Questioner initiates a refinement process featuring self-questioning (Socratic Questioning). Specifically, a VLM $\mathcal{Q}$, acting as a \textit{socratic questioner}, queries itself about omissions or inaccuracies within $X_{gold}$ in order to direct its attention to relevant visual regions. As a result, the discovered valuable cues are formatted into Q\&A pairs $X_{socra}$.
\begin{equation}
    X_{socra}=\{(X_{q,j}, X_{a,j})\}_{j=1}^M=\mathcal{Q}(X_v, X_{gold} \mid y=0)
\end{equation}
where the number of Q\&A pairs $M$ varies by case. The training data $\{(X_v, X^{ann}_{gold}), X_{socra}^{ann}\}$, with associated $y^{ann}=0$, for $\mathcal{Q}$ is obtained by 
\begin{equation}
    X^{ann}_{socra} = \mathcal{A}(X_v,X^{ann}_{gold} \mid y^{ann}=0)
\end{equation}
where $X^{ann}_{socra}$ comprises annotated Q\&A pairs. And the training objective $\mathcal{O}_{socra}$ maximizes the likelihood of producing each annotated Q\&A pair $X^{ann}_{socra,j} = (X^{ann}_{q,j}, X^{ann}_{a,j})$ given $X_v$ and all previous rounds $X^{ann}_{socra,<j} = (X^{ann}_{q,<j}, X^{ann}_{a,<j})$:
\begin{equation} 
\begin{split} 
\mathcal{O}_{socra} &= \sum_{j=1}^{M} \bigg[ \log P(X_{a,j}^{ann} \mid X_v, X^{ann}_{gold}, X_{q,j}^{ann}) + \\ 
&\quad \log P(X^{ann}_{q,j} \mid X_v, X^{ann}_{gold}, X_{q,<j}^{ann}, X_{a, <j}^{ann}) \bigg] 
\end{split} 
\label{eq:sq_ob} 
\end{equation}

Next, $X_{socra}$ augments $X_{gold}$ to generate the final, improved description $X_{des}$: (Example Q\&A pairs and final description are shown in \cref{fig:exam}). 
\begin{equation}
    X_{des} = \mathcal{D}(X_v, X_{gold}, X_{socra} \mid y=0)
\end{equation}
The training data for $\mathcal{D}$ is $\{(X_v, X^{ann}_{gold}, X_{socra}^{ann}),X^{ann}_{des}\}$ where $y^{ann}=0$ and the annotated description $X_{des}^{ann}$ is obtained via, 
\begin{equation}
    X_{des}^{ann} = \mathcal{A}(X_v,X^{ann}_{gold},X^{ann}_{socra} \mid y^{ann}=0) 
\end{equation}
Considering both possibilities of validity, the training objective $\mathcal{O}_{des}$ maximizes the likelihood of generating annotated description $X^{ann}_{des}$ given verified intermediate results:
\begin{equation}
\begin{split}
   \mathcal{O}_{des}=\log P(X_{des}^{ann} \mid X_v, X^{ann}_{gold})y^{ann} + \\
 \log P(X^{ann}_{des} \mid X_v, X^{ann}_{gold}, X_{socra}^{ann})(1-y^{ann})
\end{split}
\label{eq:dec_ob}
\end{equation}

\subsection{Semantic Entailment Evaluator}

To quantitatively assess how precisely a VLM's open-ended description $X_{des}$ portrays the actions, intentions, and context of individuals in visual inputs, the Output Evaluator $\mathcal{E}$ within our Semantic Entailment Evaluator module \cref{fig:frame}~(b) ascertains the factual consistency between $X_{des}$ and a set of ground truth key details $X^{GT}_{gold}=\{X_{gold,i}\}_{i=1}^K$ (termed \textbf{Ground Truth Gold Points}). Unlike the annotation of $X^{ann}_{gold}$, we provide $\mathcal{A}$ with GT action label $B$ as a strong prior when generating $X^{GT}_{gold}$ to ensure fidelity ($X_{gold,K}$ explicitly contains $B$).
For each ground truth point in $X^{GT}_{gold}$, $\mathcal{E}$ classifies its semantic relationship with $X_{des}$ as entailment [2], neutral [1], or contradiction [0]:
\begin{equation}
   R = \mathcal{E}(X_{des}, X^{GT}_{gold}) 
\end{equation}
where $R=\{R_i\}_{i=1}^K\in \{0,1,2\}^K$ represents the determined entailment relations. A demonstration of semantic entailment evaluation outcomes is given in \cref{fig:see}. 

From these relations, we compute the Semantic Entailment Score $S_{\rm sem\_ent} \in [0,1]$:
\begin{equation}
S_{sem\_ent} = \frac{1}{2} \left( \frac{N_e - N_c}{K} + 1 \right)
\label{eq:ent_score}
\end{equation}

where $N_e$ and $N_c$ denote the counts of entailments and contradictions in $R$, respectively. The initial score $S_{\rm ent\_init} \in [-1,1]$ is normalized to obtain $S_{\rm sem\_ent}$. This score serves as an objective metric for semantic consistency and will measure VLMs' action understanding performance in \cref{experiment}.

Since a single VLM performs all roles—gold points extractor $\mathcal{G}$, validator $\mathcal{V}$, questioner $\mathcal{Q}$, and describer $\mathcal{D}$ in this work—we combine all objectives for unified fine-tuning:
\begin{equation}
    \mathcal{O}_{overall} = \mathcal{O}_{GPE} + \mathcal{O}_{valid} + \mathcal{O}_{socra} + \mathcal{O}_{des}
\end{equation}
This comprehensive objective enhances the VLM's fine-grained reasoning capabilities across all module functions.

\section{Experiment}
\label{experiment}
% Our experiments are designed to validate the efficacy of the GPS framework by addressing three key research questions: 1. Does the GPS framework enhance lightweight VLMs for precise human action articulation, and critically, does this enhancement generalize to new actions and scenes in a zero-shot setting? 2. Is the core training strategy in GPS crucial for enhancing VLM performance, or can its effects be replicated by CoT prompt engineering? 3. What is the specific contribution of each GPS reasoning stage—extraction, validation, and refinement—to the VLM's final description quality?

We conduct extensive experiments across a range of lightweight VLMs to demonstrate the advantages of our GPS framework. Our experiments aim to investigate two key aspects of the GPS framework: (i) the performance boost when incorporating the GPS framework on VLMs for fine-grained human action understanding tasks and the adaptability to unseen scenarios, and (ii) the necessity of the GPS framework (\eg, whether it could be replaced by Chain-of-Thought prompt engineering). Additionally, we conduct an ablation study to analyze the contribution of each core module in the GPS framework, providing a thorough understanding of our proposed approach.

\subsection{Experiment Setup}
\label{setup}
\textbf{Benchmark}: We construct an \textit{instruction-tuning dataset} and an \textit{assessment benchmark} using CAP~\cite{byrne2023fine}, a corpus designed for fine-grained human action recognition. To be helpful for fine-grained action understanding, all selected action categories feature rich subject interactions. Specifically, the instruction-tuning dataset consists of 14,281 video frames from 80 action categories. For training, it provides video frames $X_v$, annotated gold points $X_{gold}^{ann}$, ground-truth gold points $X_{gold}^{GT}$, validation $y^{ann}$, optional socratic Q\&A pairs $X^{ann}_{socra}$ and final description $X^{ann}_{des}$. In contrast, the assessment benchmark is designed purely for evaluation and contains 3,321 video frames from 40 categories, comprising only video frames $X_v$ and their ground truth gold points $X_{gold}^{GT}$. To evaluate generalization, the assessment benchmark is further divided into a held-in set (20 categories, 1,117 frames) and a held-out set (20 categories, 2,204 frames). The action categories in the held-in set are seen during instruction tuning, but the video frames are from different scenarios. Conversely, the categories in the held-out set are absent from the tuning set, enabling zero-shot evaluation.

\textbf{Models}: We select four SOTA lightweight LVLMs: Qwen-VL-Chat~\cite{bai2023qwenvlversatilevisionlanguagemodel}, MiniGPT-v2~\cite{chen2023minigpt}, LLaMA3-LLaVA-NeXT-8B (LLaVA-NeXT-8B)~\cite{li2024llavanext-strong} and LLaVA-OneVision-Qwen2-7B (LLaVA-OV-7B)~\cite{lillava} for our experiments and VLMs' fine-tunings are performed on four A800-SXM4-80GB. In addition to these lightweight models, we also evaluate a SOTA open-source large VLM, LLaVA-OneVision-Qwen2-72B-ov-chat (LLaVA-OV-72B)\cite{lillava}, and the proprietary model, GPT-4o\cite{hurst2024gpt}, on our benchmark. This serves a dual purpose: first, to assess the capabilities of leading large-scale VLMs on our specialized task of fine-grained human action understanding; and second, to compare their performances against our enhanced lightweight models to highlight the effects of improvements.

\textbf{Finetuning \& Prompting}: The GPS framework, as detailed in \cref{method}, is built upon a four-stage reasoning structure: gold points extraction, gold points validation, selective questioning refinement, and fine-grained description. The application of this framework consists of two distinct phases: In the \textbf{Finetuning phase}, the instruction-tuning dataset is used to explicitly train the VLMs on how to perform each individual stage of the reasoning structure. Subsequently, in the \textbf{Prompting phase}, the VLM is prompted to replicate this trained, multi-stage process in inference. Critically, during inference necessary inputs like gold points, validation outcome and refined Q\&A pairs are generated by the VLM in real-time and fed into its own subsequent reasoning steps via appropriate prompts.

\begin{figure*}[h]
\centering
\includegraphics[width=0.95\linewidth]{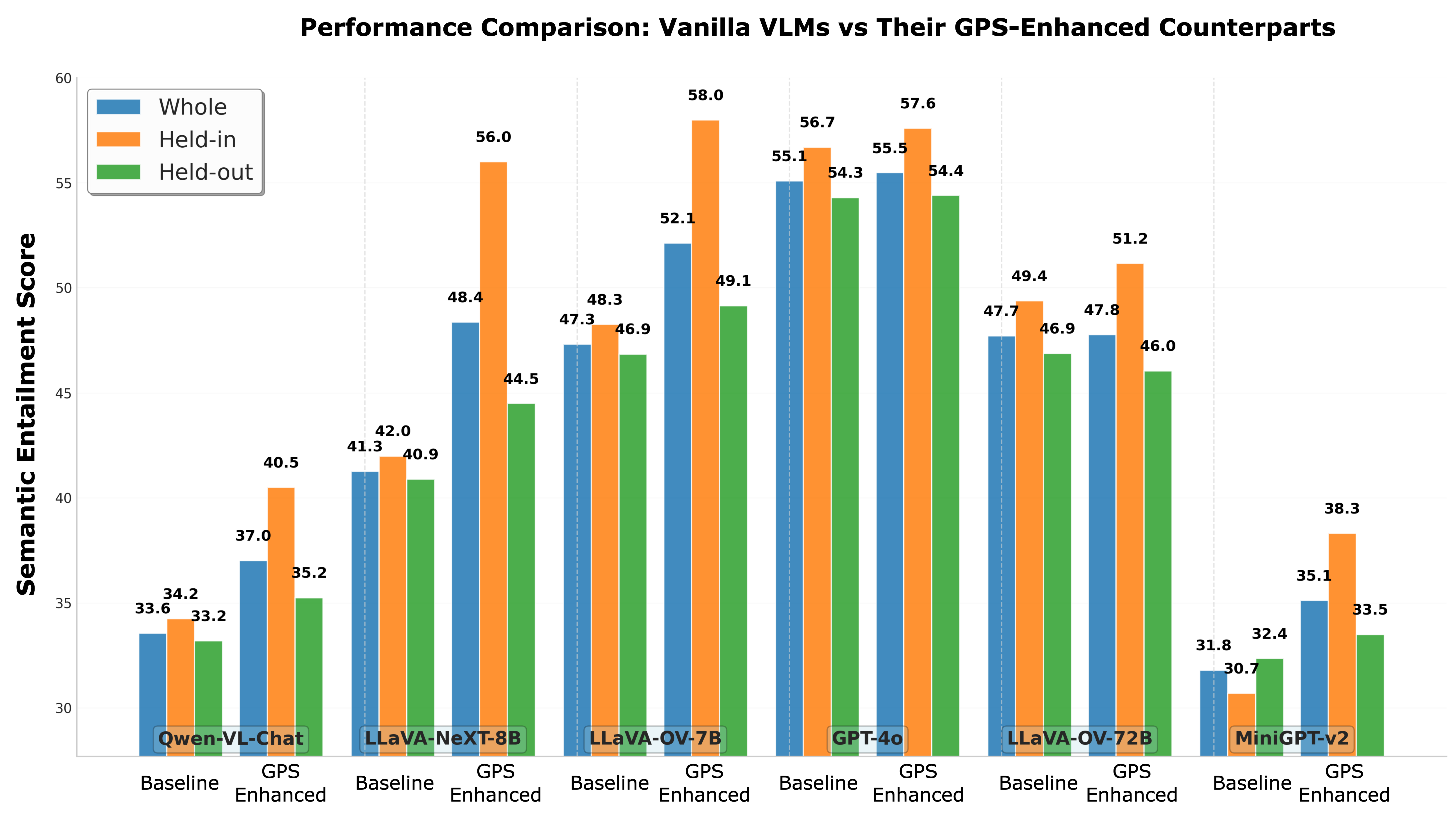}
\caption{Histogram showcasing the performances of various VLMs (vanilla \& GPS-Enhanced version) on our benchmark.} 
\label{fig:hist}
\end{figure*}

\subsection{Performance Boost with GPS}

In this experiment we evaluate to what extent, the GPS framework boost the performance of the lightweight VLMs and whether such improvements can generalize to new scenarios and new actions. To highlight the effectiveness of the GPS framework, we evaluate the performance of various vanilla VLMs against their GPS-enhanced counterparts on our benchmark, with performance measured by semantic entailment scores \ref{eq:ent_score}. The term ``GPS-Enhanced'' indicates that lightweight VLMs adopt GPS fine-tuning and prompting phases, while for LLaVA-OV-72B and GPT-4o, where fine-tuning is infeasible, this refers solely to GPS prompting. To assess learning efficacy and zero-shot generalization, we report results on the whole benchmark as well as on held-in (seen) and held-out (unseen) sets. This evaluation protocol is maintained throughout all subsequent experimental sections.

As illustrated in \cref{fig:hist}, the evaluation results align closely with human intuition that more advanced models achieve higher ratings, supporting the credibility of our Semantic Entailment Evaluator. Compared to vanilla baselines, all lightweight VLMs achieve substantially better scores on held-in, held-out, and whole benchmark after GPS enhancement. Notably, GPS-enhanced LLaVA-NeXT-8B and LLaVA-OV-7B demonstrate performance comparable to GPT-4o on the held-in set, indicating that GPS effectively enhances lightweight VLMs' fine-grained action understanding capabilities, producing information-dense and factually accurate outputs. For the already powerful GPT-4o and LLaVA-OV-72B, the effect of this multi-stage prompting is less pronounced. LLaVA-OV-72B's minor performance dip can be attributed to instability from applying complex prompt engineering to a large model for such a complicated task.

The substantial improvements on the held-in set demonstrate that GPS effectively teaches VLMs to understand actions encountered during training. More importantly, despite smaller improvement margins, the consistent enhancements on the held-out set validate the zero-shot generalization capability of GPS-Enhanced VLMs across new scenarios and actions. This achievement is particularly significant given the held-out set's inherent difficulty—vanilla models typically underperform here compared to the held-in set, and action categories are entirely unseen during training. Even modest improvements on this challenging set therefore represent substantial progress. GPS does not merely help models memorize training examples but teaches a generalizable analytical framework: observe key details, verify accuracy, and identify missing information. Since this analytical skill transcends specific tasks, it generalizes effectively to new scenarios and completely unseen actions.

\begin{table}[h]
\centering
\begin{tcolorbox}[
    colback=gray!10,
    colframe=black,
    width=0.95\linewidth,
    arc=1mm,
    auto outer arc,
    boxrule=0.5pt,
]
Please offer a visually grounded summarizing description of the ongoing human action, as well as the associated intention and context. Reasoning supported by reliable external knowledge and visual details is permitted. Also be sure to leverage the temporal changes in the frames and state the recognized action (including the interacted objects if possible) with one sentence at last.
\end{tcolorbox}
\caption{Generic prompt used as our baseline prompt.}
\label{tab:gen_prom}
\end{table}
\subsection{Necessity of GPS Framework}
\begin{table}[h] 
    \centering  
    %\footnotesize
    \small
    \setlength{\tabcolsep}{3pt}  % 调整列间距  
    \renewcommand{\arraystretch}{1.2} 
    \resizebox{\columnwidth}{!}{
    \begin{tabular}{@{}p{3.2cm}<{\raggedright}|lll@{}}  
    \toprule  
    \multirow{2}{*}{\textbf{Models}} &    
    \multicolumn{3}{c}{\textbf{Semantic Entailment Score}} \\
    \cmidrule(lr){2-4}  
    & \textbf{Held-in} & \textbf{Held-out} & \textbf{Whole}  \\
    \midrule  
    \textbf{Qwen-VL-Chat}   
    & 34.25 & 33.20  & 33.56 \\
    \textbf{+ Prompting} &   
    33.40 {\color{degradered}($\downarrow$2.48\%)} &   
    31.40 {\color{degradered}($\downarrow$5.42\%)} &      
    32.08 {\color{degradered}($\downarrow$4.41\%)} \\
    \textbf{+ Finetuning} &   
    34.90 {\color{improvegreen}($\uparrow$1.90\%)} &   
    33.65 {\color{improvegreen}($\uparrow$1.36\%)} &     
    34.07 {\color{improvegreen}($\uparrow$1.52\%)} \\
    \textbf{+ Combined} &    
    40.50 {\color{improvegreen}($\uparrow$18.25\%)} &   
    35.25 {\color{improvegreen}($\uparrow$6.17\%)} &      
    37.02 {\color{improvegreen}($\uparrow$10.31\%)} \\

    \midrule  
    \textbf{MiniGPT-v2}   
    & 30.70 & 32.35 & 31.80 \\
    \textbf{+ Prompting} &   
    29.15 {\color{degradered}($\downarrow$5.05\%)} &   
    31.00 {\color{degradered}($\downarrow$4.17\%)} &     
    30.38 {\color{degradered}($\downarrow$4.47\%)} \\
    \textbf{+ Finetuning} &   
    37.00 {\color{improvegreen}($\uparrow$20.52\%)} &   
    32.50 {\color{improvegreen}($\uparrow$0.46\%)} &     
    34.02 {\color{improvegreen}($\uparrow$6.98\%)} \\
    \textbf{+ Combined} &    
    38.32 {\color{improvegreen}($\uparrow$24.82\%)} &   
    33.50 {\color{improvegreen}($\uparrow$3.55\%)} &   
    35.12 {\color{improvegreen}($\uparrow$10.44\%)} \\

    \midrule  
    \textbf{LLaVA-NeXT-8B}   
    & 42.00 & 40.90 & 41.27 \\
    \textbf{+ Prompting} &   
    41.25 {\color{degradered}($\downarrow$1.79\%)} &   
    37.10 {\color{degradered}($\downarrow$9.29\%)} &    
    38.50 {\color{degradered}($\downarrow$6.71\%)} \\
    \textbf{+ Finetuning} &   
    54.40 {\color{improvegreen}($\uparrow$29.52\%)} &   
    42.80 {\color{improvegreen}($\uparrow$4.65\%)} &   
    46.70 {\color{improvegreen}($\uparrow$13.16\%)} \\
    \textbf{+ Combined} &    
    56.00 {\color{improvegreen}($\uparrow$33.33\%)} &   
    44.50 {\color{improvegreen}($\uparrow$8.80\%)} &      
    48.37 {\color{improvegreen}($\uparrow$17.20\%)} \\
    % \textbf{+ 14k full} & 
    % 53.75 {\color{improvegreen}($\uparrow$27.98\%)} & 
    % 41.15 {\color{improvegreen}($\uparrow$0.61\%)} & 
    % 45.39 {\color{improvegreen}($\uparrow$9.98\%)} \\
    % \textbf{+ 14k no prop} & 
    % 53.35 {\color{improvegreen}($\uparrow$27.02\%)} & 
    % 42.15 {\color{improvegreen}($\uparrow$3.06\%)} & 
    % 45.92 {\color{improvegreen}($\uparrow$11.27\%)} \\
    % \textbf{+ 14k no qa} & 
    % 56.80 {\color{improvegreen}($\uparrow$35.24\%)} & 
    % 43.80 {\color{improvegreen}($\uparrow$7.09\%)} & 
    % 48.17 {\color{improvegreen}($\uparrow$16.72\%)} \\
    
    \midrule  
    \textbf{LLaVA-OV-7B}   
    & 48.26 & 46.85 & 47.32 \\
    \textbf{+ Prompting} &   
    50.23 {\color{improvegreen}($\uparrow$4.08\%)} &   
    46.30 {\color{degradered}($\downarrow$1.17\%)} &    
    48.68 {\color{improvegreen}($\uparrow$2.87\%)} \\
    \textbf{+ Finetuning} &   
    54.90 {\color{improvegreen}($\uparrow$13.76\%)} &   
    47.90 {\color{improvegreen}($\uparrow$2.24\%)}  &     
    49.20 {\color{improvegreen}($\uparrow$3.97\%)} \\
    \textbf{+ Combined} &    
    58.00 {\color{improvegreen}($\uparrow$20.18\%)} &   
    49.15 {\color{improvegreen}($\uparrow$4.91\%)} &   
    52.13 {\color{improvegreen}($\uparrow$10.16\%)} \\
    % \textbf{+ 14k full} & 
    % 54.69 {\color{improvegreen}($\uparrow$13.32\%)} & 
    % 45.55 {\color{degradered}($\downarrow$2.77\%)} & 
    % 48.63 {\color{improvegreen}($\uparrow$2.77\%)} \\
    % \textbf{+ 14k no prop} & 
    % 54.44 {\color{improvegreen}($\uparrow$12.81\%)} & 
    % 45.52 {\color{degradered}($\downarrow$2.84\%)} & 
    % 48.52 {\color{improvegreen}($\uparrow$2.54\%)} \\
    % \textbf{+ 14k no qa} & 
    % 57.38 {\color{improvegreen}($\uparrow$18.90\%)} & 
    % 48.60 {\color{improvegreen}($\uparrow$3.74\%)} & 
    % 51.55 {\color{improvegreen}($\uparrow$8.94\%)} \\
    
    \midrule  
    \textbf{LLaVA-OV-72B}
    & 49.38 & 46.88 & 47.72 \\
    \textbf{+ Prompting} &   
    51.16 {\color{improvegreen}($\uparrow$3.60\%)} &   
    46.05 {\color{degradered}($\downarrow$1.77\%)} &   
    47.77 {\color{improvegreen}($\uparrow$0.10\%)} \\
    
    \midrule  
    \textbf{GPT-4o}   
    & 56.70 & 54.30 & 55.10 \\
    \textbf{+ Prompting} &   
    57.60 {\color{improvegreen}($\uparrow$1.59\%)} &   
    54.40 {\color{improvegreen}($\uparrow$0.18\%)} &      
    55.48 {\color{improvegreen}($\uparrow$0.69\%)} \\

    \bottomrule  
    \end{tabular}  
    }
    \caption{Assessing the independent and synergistic effects of GPS finetuning and prompting phases described in \cref{setup}.}
    \label{tab:ftpt}  
\end{table}

In this experiment, we investigate the necessity of the GPS framework by analyzing whether its performance can be achieved through pure Chain-of-Thought prompt engineering. To validate this necessity, we isolate the prompting phase of GPS—itself a form of CoT prompt engineering—from the overall framework to assess its individual effect relative to the finetuning phase. To this end, we designed the following four distinct experimental configurations: The \textbf{baseline configuration} employs a generic prompt \cref{tab:gen_prom} to guide a vanilla VLM in action description. The \textbf{+ Prompting} configuration involves only GPS prompting, directing the vanilla VLM to follow GPS's four-stage reasoning structure. The \textbf{+ Finetuning} configuration consists solely of finetuning, where the VLM learns GPS reasoning stages but receives only generic prompting \cref{tab:gen_prom} at inference. Finally, the \textbf{+ Combined} setting represents complete GPS Enhancement, first training the VLM on four-stage reasoning during finetuning, then prompting the trained model to replicate the full reasoning process during inference.

Results in \cref{tab:ftpt} show that \textbf{+ Prompting} caused performance degradation for nearly all lightweight VLMs, with only minor improvement in LLaVA-OV-7B. This reveals a key insight: complex, multi-stage prompts become cognitive burdens rather than helpful guidance for lightweight VLMs. Lacking inherent capability to follow four-stage reasoning, these models likely generate flawed Gold Points or fail at self-verification, causing the entire reasoning chain to collapse. This demonstrates that complex reasoning skills must be learned through targeted training rather than elicited by inference-time prompts.
In contrast, \textbf{+ Finetuning} led to substantial performance enhancements across all models, confirming that our training strategy is essential for equipping models with necessary skills. Peak performance was achieved exclusively under the \textbf{+ Combined} setting, revealing crucial synergy between training and prompting in our framework. The finetuning phase endows models with GPS reasoning capabilities, while the prompting phase provides structured guidance during inference, ensuring these learned abilities are reliably invoked in correct sequence.

\subsection{Ablation Study of Core Components}
\begin{table}[h]
\centering
\footnotesize 
\setlength{\tabcolsep}{4pt} % <--- 添加此行来缩小列间距
\begin{tabular}{|p{2.1cm}|p{1.8cm}|p{1.8cm}|p{1.8cm}|}
\hline
LLaVA-NeXT-8B & Held-in & Held-out & Whole \\
\hline
w/o GPE & 53.35 {\color{degradered}($\downarrow$4.73\%)} & 42.15 {\color{degradered}($\downarrow$5.28\%)} & 45.92 {\color{degradered}($\downarrow$5.07\%)} \\
w/o GPV & 53.75 {\color{degradered}($\downarrow$4.02\%)} & 41.15 {\color{degradered}($\downarrow$7.53\%)} & 45.39 {\color{degradered}($\downarrow$6.16\%)} \\
w/o SQR & 56.80 {\color{improvegreen}($\uparrow$1.43\%)} & 43.80 {\color{degradered}($\downarrow$1.57\%)} & 48.17 {\color{degradered}($\downarrow$0.41\%)} \\
full GPS & 56.00 & 44.50 & 48.37 \\
\hline
\end{tabular}

\vspace{0.2cm}

\begin{tabular}{|p{2.1cm}|p{1.8cm}|p{1.8cm}|p{1.8cm}|}
\hline
LLaVA-OV-7B & Held-in & Held-out & Whole \\
\hline
w/o GPE & 54.44 {\color{degradered}($\downarrow$6.14\%)} & 45.52 {\color{degradered}($\downarrow$7.39\%)} & 48.52 {\color{degradered}($\downarrow$6.93\%)} \\
w/o GPV & 54.69 {\color{degradered}($\downarrow$5.71\%)} & 45.55 {\color{degradered}($\downarrow$7.32\%)} & 48.63 {\color{degradered}($\downarrow$6.71\%)} \\
w/o SQR & 57.38 {\color{degradered}($\downarrow$1.07\%)} & 48.60 {\color{degradered}($\downarrow$1.12\%)} & 51.55 {\color{degradered}($\downarrow$1.11\%)} \\
full GPS & 58.00 & 49.15 & 52.13 \\
\hline
\end{tabular}
\caption{Ablations on prompting LLaVA-NeXT-8B and LLaVA-OV-7B (both GPS finetuned). Three ablated components are Gold Points Extraction (GPE), Gold Points Validation (GPV) and Selective Questioning Refinement (SQR).}
\label{tab:abla}
\end{table}

To evaluate each module's contribution in the GPS framework, we conduct a prompting ablation study on LLaVA-NeXT-8B and LLaVA-OV-7B models after GPS finetuning. During inference, we systematically ablate each of the three steps--GPE, GPV, and SQR (as shown in \cref{tab:abla})---from the prompting-based reasoning process while retaining the other two. We perform this ablation on finetuned VLMs because only models capable of properly executing each GPS reasoning stage can provide faithful assessment of component effectiveness. We examine three ablation settings: (i) \textbf{Without GPE}: The VLM uses only visual input to choose between direct description generation and self-questioning for additional visual clues. (ii) \textbf{Without GPV}: The VLM skips validation of generated gold points and proceeds directly to selective refinement, but lacks validation feedback to guide this selection. (iii) \textbf{Without SQR}: The VLM generates the final description based solely on generated gold points and their validation results without further refinement.

The ablation results in \cref{tab:abla} confirm that all components contribute meaningfully while revealing their distinct functional roles. Gold Points Extraction (GPE) and Gold Points Validation (GPV) serve as foundational components, with their removal causing the most substantial performance drops, demonstrating that the reasoning chain depends critically on accurately extracting and verifying core factual information. After GPE extracts facts and GPV validates them, SQR identifies missing details and resolves ambiguities, transforming factually correct descriptions into more informative outputs. The smaller performance drop from SQR removal indicates its primary role is improving output quality rather than establishing correctness. This ablation study demonstrates that GPS effectively balances informational richness and factual fidelity by first ensuring accurate fact extraction and validation, then systematically enhancing descriptive detail without compromising correctness.

\subsection{Qualitative Evaluation}
\begin{figure}[t]
\centering  
\includegraphics[width=\linewidth]{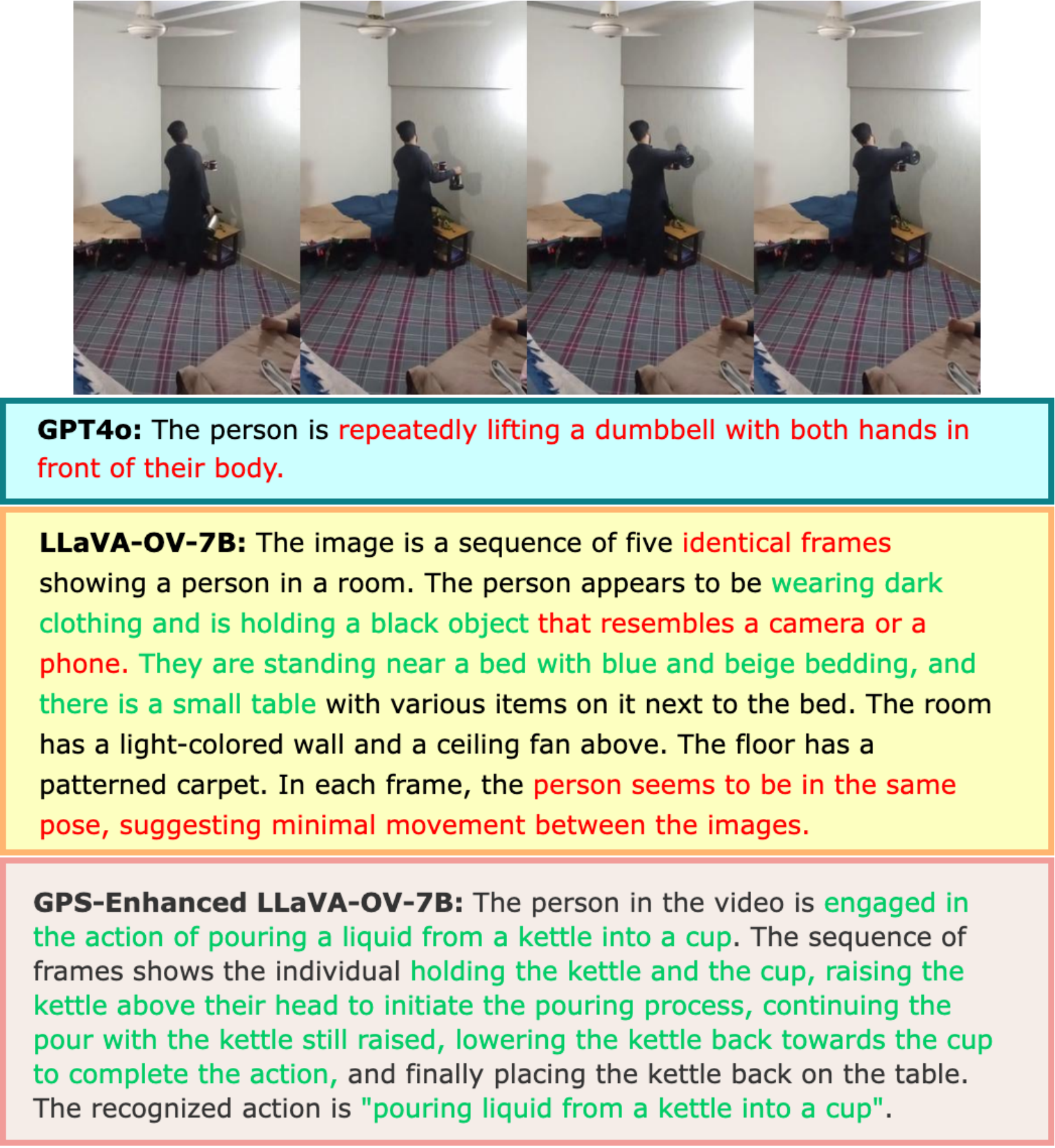}
\caption{A qualitative example, with GT label ``person pours coffee into mug'',  demonstrating the usage outcomes of vanilla GPT-4o, vanilla LLaVA-OV-7B and GPS-Enhanced LLaVA-OV-7B on a video frames from our benchmark.}
\label{fig:qual}
\end{figure}

\cref{fig:qual} presents qualitative results of GPS-Enhanced human action understanding, demonstrating the effectiveness of the GPS framework. 
The baseline models show critical failures in fine-grained action understanding despite adequate general scene description. GPT-4o hallucinates both object and action, misinterpreting pouring from a kettle as ``lifting a dumbbell'', while vanilla LLaVA-OV-7B correctly perceives the static environment but fails to grasp temporal dynamics, incorrectly concluding the frames are ``identical'' with ``minimal movement''.
In contrast, the GPS-Enhanced model accurately identifies and sequences the core action. Rather than providing static descriptions, it correctly decomposes the event into constituent steps: holding objects, raising the kettle, pouring, and lowering it back. This demonstrates that GPS enables VLMs to move beyond coarse scene-level perception to structured, step-by-step reasoning that accurately captures fine-grained action semantics.

% \cref{fig:qual} provides a compelling qualitative demonstration of the GPS framework's efficacy. The baseline models, while capable of general scene description, exhibit critical failures in fine-grained action understanding. For instance, GPT-4o hallucinates both the object and the action, misinterpreting pouring from a kettle as "lifting a dumbbell." Similarly, the vanilla LLaVA-OV-7B correctly perceives the static environment but fails to grasp the temporal dynamics, incorrectly concluding the frames are ``identical'' with ``minimal movement'' and misidentifying the key object.

% In stark contrast, the GPS-Enhanced model transcends these limitations. It not only retains the ability for holistic scene comprehension but also precisely identifies and sequences the core action. Instead of a static description, it correctly decomposes the event into its constituent steps: holding the objects, raising the kettle, pouring, and lowering it back. This demonstrates that GPS endows the VLM with the ability to move beyond coarse scene-level perception to a structured, step-by-step reasoning process, allowing it to accurately capture the fine-grained semantics of human actions.

\section{Conclusion}
We introduced the GPS framework to enable fine-grained human action understanding for domestic robots through self-guided multimodal reasoning in lightweight VLMs. Our three-module approach—Gold Points Extraction, Selective Socratic Questioning, and Semantic Entailment Evaluation—resolves the trade-off between informational richness and factual fidelity in vision-language models. Experiments demonstrate that GPS-enhanced lightweight VLMs achieve substantial improvements across seen and unseen action categories, with some reaching performance comparable to larger proprietary systems while maintaining superior factual accuracy. The framework's generalization to new scenarios validates its real-world applicability, establishing a systematic approach for information-dense yet factually grounded action descriptions that enable trustworthy human-robot interaction in domestic environments.

\section{Acknowledgement}
This work was supported by National Science and Technology Major Project (No. 2022ZD0114903) and Beijing Natural Science Foundation (funding Number L258013).

\balance
\setstretch{0.95}
\bibliographystyle{ieeetr}
\bibliography{reference_header,reference}

\end{document}